# Exploring Personalized Federated Learning Architectures for Violence Detection in Surveillance Videos

Mohammad Kassir
Mathematics Department
CY Tech
Paris, France
kassirmoha@cy-tech.fr

Siba Haidar
*Learning, Data and Robotics (LDR)*
*ESIEA Lab, ESIEA*
Paris, France
siba.haidar@esiea.fr

Antoun Yaacoub
*Learning, Data and Robotics (LDR)*
*ESIEA Lab, ESIEA*
Paris, France
antoun.yaacoub@esiea.fr

***Abstract*— The challenge of detecting violent incidents in urban surveillance systems is compounded by the voluminous and diverse nature of video data. This paper presents a targeted approach using Personalized Federated Learning (PFL) to address these issues, specifically employing the Federated Learning with Personalization Layers method within the Flower framework. Our methodology adapts learning models to the unique data characteristics of each surveillance node, effectively managing the heterogeneous and non-IID nature of surveillance video data. Through rigorous experiments conducted on balanced and imbalanced datasets, our PFL models demonstrated enhanced accuracy and efficiency, achieving up to 94.7% on balanced data and above 91% across heterogeneous scenarios. This study underscores the potential of PFL to significantly improve the scalability and effectiveness of surveillance systems, offering a robust, privacy-preserving solution for violence detection in complex urban environments.**

***Keywords*— *Personalized Federated Learning, Violence Detection, Surveillance Systems, Model-Agnostic Meta-Learning, Personalization Layers, Hybrid Learning Models, Non-IID Data***

## I. Introduction

Surveillance systems in urban environments are critical for ensuring public safety, as they are tasked with the challenging role of accurately detecting violent incidents in real-time. These systems must handle vast amounts of complex video data, which presents significant analytical challenges that surpass those encountered with static images or text data. Traditional federated learning (FL) models often fall short in these scenarios due to their inability to effectively manage the dynamic and detailed nature of video footage.

This paper explores the application of Personalized Federated Learning (PFL) to advance violence detection capabilities in urban surveillance systems. PFL offers a compelling solution by enabling customized model training across decentralized nodes, each adapting to localized data peculiarities inherent in urban surveillance footage. This approach allows for the effective handling of diverse, non-IID (independently and identically distributed) data characteristics typical of such environments, thereby enhancing both the accuracy and efficiency of detection systems. Furthermore, PFL addresses the substantial computational demands of processing high-volume video data, optimizing the use of bandwidth and storage without compromising privacy or system performance.

By delving into various PFL architectures, this paper aims to demonstrate how tailored learning processes can significantly improve the precision and reliability of violence detection in surveillance videos. Through rigorous methodologies and detailed experiments, we explore the potential of PFL to transform surveillance practices, making public spaces safer while adhering to stringent data privacy standards. This work, inspired from the foundational research [1], emphasizes the importance of developing systems that not only detect but also interpret and describe violent interactions, thereby enhancing situational awareness and response capabilities.

The structure of the paper is organized as follows. Section II provides a comprehensive background on the computational challenges of analyzing surveillance video for violence detection and the historical methods used in this field. Section III delves into the state-of-the-art techniques in violence detection and federated learning, introducing the concept of Personalized Federated Learning (PFL) and its relevant methodologies, including Model-Agnostic Meta-Learning (MAML) and Parameter Decoupling. Section IV outlines the methodology adopted for implementing PFL, describing model replication, data processing, and integration of personalization layers. Section V details the experimental setup, datasets used, configurations, and the results obtained from various experiments, followed by a comparative analysis of the model's performance. Finally, Section VI concludes the paper, summarizing the key findings, discussing the implications of the results, and suggesting directions for future research.

## II. Background

The analysis of surveillance video for violence detection presents unique computational challenges due to the high dimensionality and temporal variability of video data. Federated Learning (FL) has emerged as a vital approach to address these demands, facilitating model training across dispersed nodes without necessitating data centralization. This methodology inherently supports privacy and bandwidth optimization but struggles with data heterogeneity across diverse surveillance settings.

Historically, violence detection, a subset of action recognition, has relied on classical methods utilizing hand-crafted features. Examples include ViF (violent flow), which detects variations in optical flow, and OViF (oriented violent flow) [2], which builds upon ViF by better exploiting orientation information within optical flows.

Optical flow refers to the perceived motion between consecutive frames in a video.

Recent advancements in deep learning have significantly improved violence detection, offering enhanced accuracy and reduced computational demands. Several architectures have been explored, such as 3D Convolutional Neural Networks (3D CNNs), which perform convolutions over video frames. Enhancements, including the integration of DenseNet concepts, have optimized these models by improving performance and reducing parameters. Additionally, combining convolutional networks with Long Short-Term Memory (LSTM) cells has been effective, as LSTM networks aggregate temporal features extracted by CNNs from video frames. Another innovative technique involves using dual input channels—one for video frames and another for optical flow—enabling models to focus on regions exhibiting motion and potential violence.

Our prior research, detailed in [3], has shown that tailored federated learning architectures can significantly balance accuracy and training efficiency in violence detection systems. This foundational work informs the advancements presented in this paper, which builds on these methodologies to further refine the application of PFL in surveillance contexts.

## III. STATE OF THE ART

In this section, we explore two important topics in the field of data science: violence detection and federated learning.

### A. Violence detection

The field of violence detection, a subset of action recognition, focuses on identifying specific actions within video footage. Historically, this area has leveraged classical methods that utilize hand-crafted features. For example, ViF (violent flow) detects variations in optical flow, while OViF (oriented violent flow) [2] builds upon ViF by better exploiting orientation information within optical flows. Optical flow refers to the perceived motion between consecutive frames in a video.

In recent years, the advent of deep learning has significantly advanced violence detection, offering improved accuracy and reduced computational demands compared to classical techniques. Several deep learning architectures have been explored. A notable approach involves 3D Convolutional Neural Networks (3D CNNs) [4], which perform convolutions over video frames. Enhancements such as integrating DenseNet concepts [5] [6] have further optimized these models, enhancing performance while minimizing parameters.

Additionally, the combination of convolutional networks with Long Short-Term Memory (LSTM) cells [7] has been investigated. LSTM networks aggregate temporal features extracted by CNNs from video frames, as demonstrated in [8]. Another innovative technique introduced by [9] employs dual input channels—one for the video frames and another for optical flow—enabling the model to concentrate on regions exhibiting motion and potential violence.

### B. Federated Learning

Federated learning (FL) represents a paradigm shift in data privacy, allowing for decentralized model training across multiple computational nodes. This framework ensures that sensitive data remains on local devices, thus preserving user privacy and adhering to stringent data regulations. Despite its advantages, traditional FL faces significant challenges, particularly when dealing with non-IID (Independently and Identically Distributed) data across different nodes. This heterogeneity can severely degrade model performance, making it imperative to explore more adaptive learning strategies.

### C. Introduction to Personalized Federated Learning (PFL)

In addressing the challenges of federated learning, Personalized Federated Learning (PFL) has emerged as a promising solution by customizing models to specific client environments and enhancing their accuracy and relevance through the integration of local data characteristics. Our review focuses on the Flower framework, which supports diverse PFL strategies such as Model-Agnostic Meta-Learning (MAML) and Parameter Decoupling. MAML is particularly notable for its ability to quickly adapt models to new tasks with minimal data, making it ideal for data-sparse environments. We explore these methodologies to assess their efficacy in personalizing models across decentralized datasets.

### D. Model-Agnostic Meta-Learning (MAML) Approach

The first approach implements the MAML framework as detailed by [10]. The Model-Agnostic Meta-Learning (MAML) approach is particularly designed to quickly adapt models to new tasks using minimal data, making it highly suitable for environments where data is scarce. Central to MAML's methodology is the concept of training a general model on a diverse array of tasks, which can then be finely tuned with a small number of gradient updates to effectively perform on new, unseen tasks. This process not only finds an optimal starting point in the parameter space for quick convergence but also ensures that the model can adjust to task-specific nuances effectively.

In the context of federated learning, the implementation of the Personalized Federated Average (Per-FedAvg) algorithm, see Fig. 1, includes several critical steps:

1. Initialization: we begin by setting up a shared model with initial parameters that act as a foundational model for subsequent personalizations.
2. Local updates: each client independently fine-tunes this base model with their own unique datasets, thereby making the model more applicable and sensitive to local data characteristics.
3. Meta-model update: adjustments are made to the global model using a meta-learning approach, enhancing its ability to adapt to new personalization data.
4. Gradient calculation: necessary modifications to the model, informed by calculated gradients and hessians for deeper insights, are made to tailor the model further based on individual user data.
5. Model aggregation: updates from all users are amalgamated to refine the global model, ensuring that it incorporates learning from the entire network.

6. Iterative process: this cycle of updates and refinements continues until the model reaches a satisfactory level of performance and personalization.

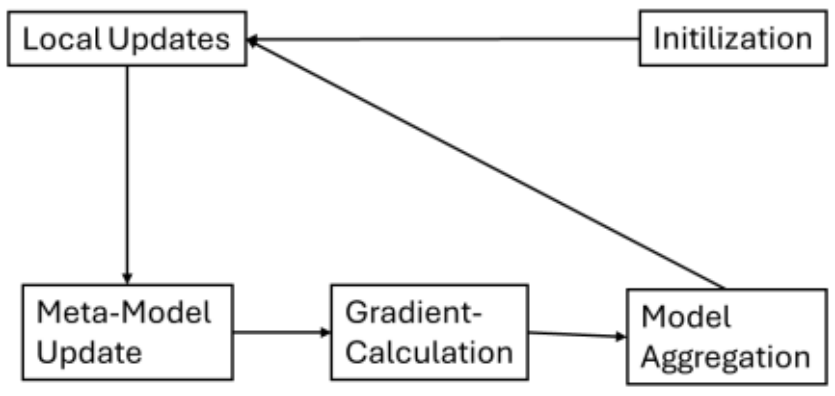

Fig. 1. PFL Process with MAML

The implementation of MAML through the Per-FedAvg algorithm in federated learning demonstrates substantial improvements in model adaptability and personalization. Utilizing local data for initial model adjustments and employing a meta-learning strategy for global updates, MAML shows a marked enhancement in managing diverse data scenarios. This approach is particularly effective in environments with sparse data availability, exhibiting superior performance in terms of convergence speed and accuracy compared to traditional federated learning methods. These findings highlight the potential of MAML in enhancing the precision and robustness of machine learning models across decentralized networks.

### E. *Personalized Layer Method (Parameter Decoupling)*

The second method, Parameter Decoupling, effectively incorporates personalized layers to adapt models within a federated learning framework. This strategy, detailed by [11], focuses on personalizing specific layers of the model while maintaining common features across tasks, striking a balance between generalization and customization. It focuses on personalizing specific layers of the model while maintaining common features across tasks, facilitating an effective balance between generalization and customization.

In practical terms, this approach is implemented through the Federated Personalization (FedPer) method, Fig. 2. FedPer restructures the machine learning model architecture by separating it into two key components: base layers, which are common and updated globally, and personalization layers, which are specific to each client and trained locally. This division allows the model to leverage shared knowledge from all participating devices, enhancing overall model performance while also allowing for significant customization based on local data peculiarities without compromising data privacy.

FedPer Client-Side Algorithm:

1. Initialization of personalization layer weights: Each client initializes its personalization layer weights randomly, establishing a personalized starting point.
2. Data count communication to server: Clients report their local data counts to the server, aiding in the coordination of the federated learning process.
3. Training loop:
    a. Reception of base layer weights: Clients receive updated base layer weights from the server, reflecting aggregated global learning.
    b. Local personalization layer updates: Using the received weights and local data, clients update their personalization layers, tailoring the model more closely to local conditions.
    c. Gradient computation and update: Clients compute gradients for the personalization layer, adjusting it based on specific learning rates.
    d. Local retention and global sharing: Updated personalization weights are retained locally, while modified base layers are sent back for global updates.

FedPer Server-Side Algorithm:

1. Base layer weight initialization: The server sets up initial base layer weights randomly, laying the groundwork for global learning.
2. Aggregation of client data inputs: The server collects data counts from each client, using this information to balance input relative to each client's data volume.
3. Distribution and aggregation cycle:
    a. Collection of updated base layers: The server receives and aggregates base layer updates from each client.
    b. Weight re-distribution: Aggregated weights are then redistributed to clients for further local training, ensuring continuous improvement and adaptation of the global model.

The effectiveness of FedPer, particularly using the Parameter Decoupling method, has been validated using datasets such as CIFAR-10 and FLICKR-AES, and through architectures like ResNet-34 and MobileNet-v1. Comparative studies highlight FedPer's superior performance over traditional federated learning (FedAvg), especially in scenarios with diverse and non-identical data distributions. The critical role of personalization layers in enhancing model efficacy underlines the method’s capability to handle statistical heterogeneity and adapt to user-specific nuances effectively.

### F. *Hybrid Approach: Meta-Learning with Personalization Layer*

The third and final approach, as discussed in [12], combines the strengths of meta-learning with personalized layers. This method employs techniques such as MAML or Meta-SGD to train foundational layers, with personalization adjustments made at the client level through dedicated layers. Known as FedMeta-Per, this approach excels by significantly improving convergence rates and personalization accuracy compared to conventional federated learning models.

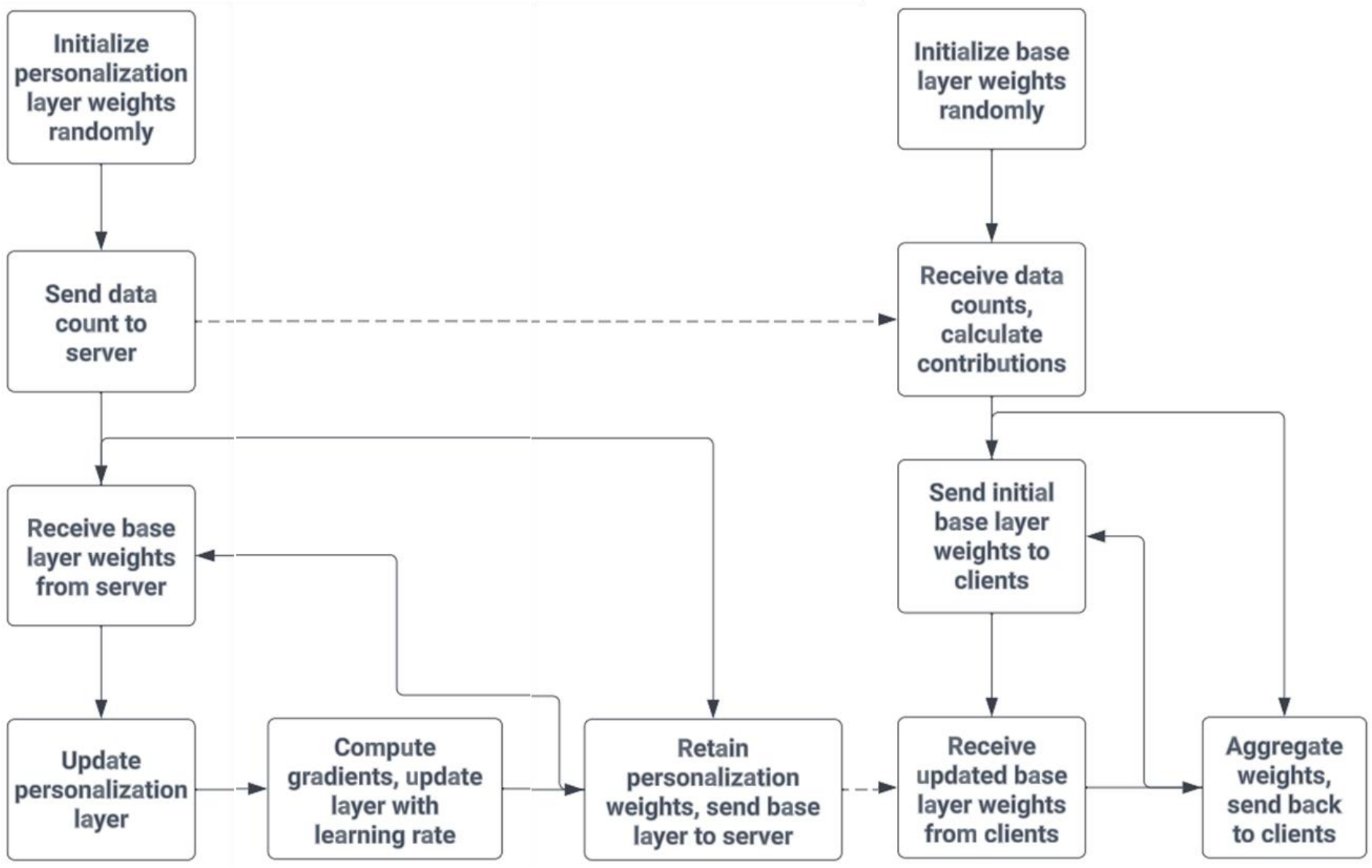

Fig. 2. Federated Learning with Personalization Layers (FedPer) Process Flow

### G. *Selection Rationale*

Our selection of the Federated Learning with Personalization Layers (FedPer) method was influenced by its recent updates and active maintenance within the Flower framework repository. This decision was strategic, favoring a method that not only demonstrates potential in theoretical advancements but is also supported by contemporary developmental efforts, ensuring relevance and applicability to current and future PFL challenges.

## IV. METHODOLOGY

Our approach leverages the Flower framework [1] to implement Personalized Federated Learning (PFL) using Model-Agnostic Meta-Learning (MAML) and Parameter Decoupling techniques. The MAML method involves training a general model on a diverse array of tasks, which is then fine-tuned with a minimal amount of data to quickly adapt to new tasks. This ensures that the model can effectively generalize and adapt to new data distributions. Parameter Decoupling involves separating the model into base and personalization layers, with the base layers trained globally and the personalization layers trained locally. This method allows the model to retain general features while adapting to specific local data characteristics. The adaptation was guided by the 'fedper' framework and draws upon methodologies outlined in [11].

### A. *Model Replication and Adaptation*

We commenced by replicating the 'Diff-Gated' model detailed in the study[3]. This model comprises a sequence of five convolutional layers followed by three fully connected layers, which effectively incorporates transfer learning to expedite and enhance the training process.

### B. *Data Processing and Model Training*

The 'Diff-Gated' model is engineered to process video data effectively by segmenting each video into 16-frame chunks, capturing essential temporal dynamics for recognizing violent activities. This involves loading videos, extracting frames, resizing them to a uniform size, and categorizing these chunks under labels such as "Fight" or "NonFight." Initially developed in Keras and TensorFlow, the model was adapted to the 'fedper' framework in PyTorch. This adaptation required key modifications, including recalibrating the padding calculations to meet PyTorch's specific requirements, which differ from those in TensorFlow. This structured dataset is crucial for training machine learning models to differentiate between fighting and non-fighting scenes, leveraging the sequential frame data for accurate classification.

### C. *Personalization Layer Integration*

For the effective integration of personalization layers, we strategically selected the split point between the convolutional and fully connected layers. This decision was informed by the understanding that convolutional layers generally serve as feature detectors, capturing general characteristics across the video data, whereas fully connected layers act as decision-making layers that adapt responses based on localized data insights. Therefore, by decoupling at this junction, we ensure that the model's feature extraction capabilities are uniformly trained across all nodes, while decision-making layers are personalized to enhance performance based on individual data distributions. Fig. 3 and Fig. 4 show a graphical representation of our model.

### D. *Objective and Rationale*

The primary objective of these methodological choices is to craft a federated learning model that not only respects the privacy constraints inherent in distributed data environments but also achieves high accuracy by leveraging local data peculiarities. By doing so, the model becomes robust enough to generalize across various surveillance contexts while being adaptable enough to offer precise violence detection tailored to specific scenarios.

[1] https://flower.ai/

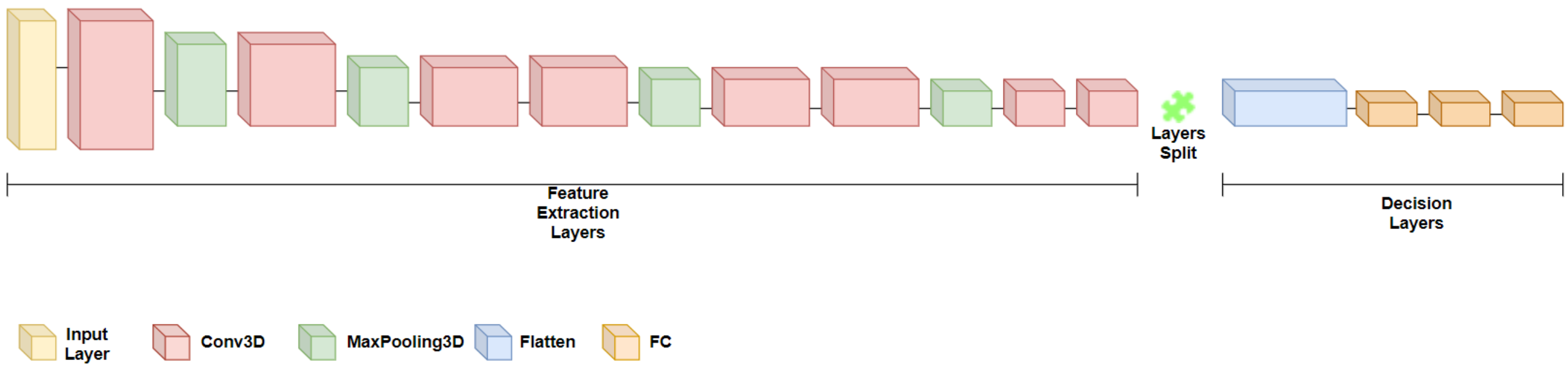

Fig. 3. Neural Network Architecture with Personalization Layer Integration for Federated Violence Detection

## V. EXPERIMENT

### A. Experimental Setup

For each configuration, client datasets were split once and fixed across all training runs. This allows us to quantify data heterogeneity per configuration based on label distribution. We conducted a series of experiments using our calculator, a high-performance computational tool optimized for intensive data analysis and machine learning tasks. This setup allowed us to evaluate our model's performance remotely and securely across both balanced and imbalanced datasets, aiming to ascertain its behavior in homogeneous and heterogeneous environments.

In each configuration, the client data splits were generated once and fixed across all training runs. To quantify inter-client heterogeneity, we computed the Jensen-Shannon (JS) divergence between client-level label distributions and the global distribution. The measured JS divergences were: Balanced = 0.0000, Imbalanced = 0.0697, Combined = 0.0147.

### B. Datasets and Distribution

Various datasets exist for violence detection, such as the movies dataset consisting of fight scenes from movies or the hockey fight dataset composed of fights from hockey games [13]. However, these datasets have specific video contexts and may not represent real-life situations. To choose the most relevant dataset, we looked for CCTV footage of real-life physical violence situations with a significant number of videos and that have been used in other studies.

The RWF-2000 dataset [9] consists of 2000 videos from various sources of CCTV cameras, with 1000 violent videos and 1000 non-violent videos. All videos are five seconds long and filmed at 30 fps. For our experiments, we used a subset of 400 videos, each divided into 16 frames to yield 3600 chunks.

The Crowd Violence dataset [14] consists of 246 videos, which were segmented similarly to provide 1270 chunks.

To quantify inter-client heterogeneity, we computed the Jensen-Shannon (JS) divergence between client-level label distributions and the global distribution for each configuration. The resulting divergences are: Balanced = 0.0000, Imbalanced = 0.0697, Combined = 0.0147.

### C. Experiment Configurations and Results

The following tables detail the configurations and results of three distinct experiments designed to evaluate our model's performance across various dataset distributions. Table 1 presents the results for the balanced dataset configuration, Table 2 for the imbalanced dataset configuration, and Table 3 shows the results for the combined dataset configuration.

TABLE I. EXPERIMENT 1 - BALANCED DATASET CONFIGURATION

| Client | Fight Samples | Non-Fight Samples | Total Samples | Loss (±std) | Accuracy (±std) |
|---|---|---|---|---|---|
| 1 | 900 | 900 | 1800 | 0.0108 ± 0.0005 | 94.6 ± 0.3% |
| 2 | 900 | 900 | 1800 | 0.0107 ± 0.0006 | 94.7 ± 0.2% |
| **Total** | 1800 | 1800 | 3600 | 0.0107 ± 0.0005 | **94.7 ± 0.2%** |

### D. Comparative Analysis

The model achieves 94.7% accuracy on balanced RWF data and maintains >91% accuracy across imbalanced and heterogeneous client scenarios. The low JS divergence observed in the Combined configuration (0.0147) suggests consistent label distributions despite variations in client data volume. This consistent performance underscores the model's robust ability to handle both balanced and imbalanced data distributions effectively. These results highlight the substantial improvements our model brings to the field, particularly in its capacity to manage and adapt to diverse data conditions.

These experiments demonstrate the robustness of our model under various data distribution scenarios. Particularly, the superior accuracy and low loss in the combined dataset configuration underline the model's effectiveness in handling heterogeneous data distributions, showcasing its potential for real-world applications.

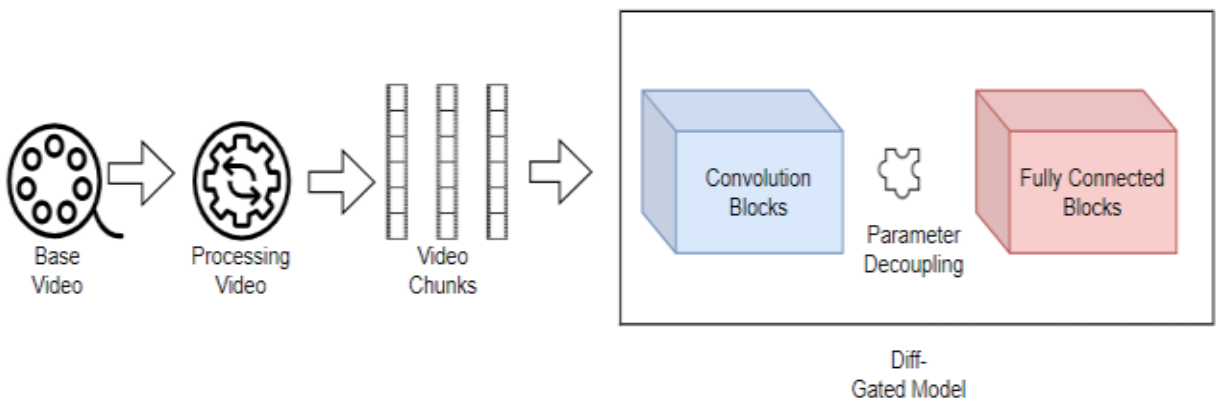

Fig. 4: Architecture of a Personalized Federated Learning Model for Violence Detection in Surveillance Videos.

TABLE II. EXPERIMENT 2 - IMBALANCED DATASET CONFIGURATION

| Client | Fight Samples | Non-Fight Samples | Total Samples | Loss (±std) | Accuracy (±std) |
|---|---|---|---|---|---|
| 1 | 641 | 27 | 668 | 0.0137 ± 0.0011 | 91.3 ± 0.7% |
| 2 | 655 | 1305 | 1960 | 0.0132 ± 0.0013 | 92.2 ± 0.8% |
| 3 | 504 | 468 | 972 | 0.0135 ± 0.0012 | 91.8 ± 0.6% |
| **Total** | 1800 | 1800 | 3600 | 0.0135 ± 0.0012 | **91.8 ± 0.7%** |

TABLE III. EXPERIMENT 3 - COMBINED DATASET CONFIGURATION

| Client | Fight Samples | Non-Fight Samples | Total Samples | Loss (±std) | Accuracy (±std) |
|---|---|---|---|---|---|
| 1 | 570 | 402 | 972 | 0.0114 ± 0.0009 | 93.1 ± 0.5% |
| 2 | 151 | 49 | 200 | 0.0122 ± 0.0010 | 92.3 ± 0.7% |
| 3 | 1019 | 777 | 1796 | 0.0110 ± 0.0008 | 93.7 ± 0.4% |
| 4 | 695 | 1207 | 1902 | 0.0115 ± 0.0007 | 93.2 ± 0.5% |
| **Total** | 2435 | 2435 | 4870 | 0.0115 ± 0.0009 | **93.1 ± 0.5%** |

TABLE IV. SUMMARY OF EXPERIMENTAL RESULTS AND CONFIGURATIONS

| Exp | configuration | # clients | JS Div. | Acc. (±std) | Loss (±std) |
|---|---|---|---|---|---|
| 1 | Balanced (RWF-2000) | 2 | 0.0000 | 94.7 ± 0.2% | 0.0107 ± 0.0005 |
| 2 | Imbalanced (RWF-2000) | 3 | 0.0697 | 91.8 ± 0.7% | 0.0135 ± 0.0012 |
| 3 | Combined (RWF + Crowd Violence) | 4 | 0.0147 | 93.1 ± 0.5% | 0.0115 ± 0.0009 |

## VI. CONCLUSION

This study investigated the application of Personalized Federated Learning (PFL) in enhancing violence detection in surveillance video systems. Our systematic exploration across several PFL architectures, notably utilizing the Flower framework for Model-Agnostic Meta-Learning (MAML), Personalized Layer Method (Parameter Decoupling), and a Hybrid Approach, demonstrates a clear advancement over traditional federated learning models. These architectures showed significant improvements in managing the heterogeneous and non-IID nature of surveillance data, thereby boosting the accuracy and efficiency of violence detection algorithms.

The experimental results affirm that PFL models can handle scalable environments across numerous clients without compromising performance. This scalability, coupled with reduced bandwidth and memory usage through localized data processing and model training, underlines the feasibility of deploying PFL in real-world surveillance settings. Notably, our model achieved an accuracy of up to 94.7% in balanced scenarios, and 93.1% in combined heterogeneous configurations, confirming its robustness to moderate non-IID distributions.

Despite these advancements, the integration of PFL into surveillance systems is not without challenges. Future research should focus on refining these models to adapt dynamically to changing environments and further improve data processing efficiency. Additionally, it is imperative to address ethical concerns and privacy issues related to AI in public surveillance to ensure responsible and acceptable use.

In summary, this research delineates a promising pathway for Personalized Federated Learning, setting a benchmark for the deployment of privacy-preserving and efficient AI technologies in the public safety domain. The findings encourage broader adoption and innovation in federated learning applications, enhancing both the technology and its implementation in critical public infrastructure.